\pdfoutput=1
\documentclass[11pt]{article}

\usepackage{acl}

\usepackage{times}  
\usepackage{latexsym}

\usepackage[T1]{fontenc}
\usepackage[utf8]{inputenc}

\usepackage{microtype}

\usepackage{inconsolata}

\usepackage{graphicx}

\usepackage{multirow}
\usepackage{colortbl}
\title{Converging Dimensions: Information Extraction and Summarization through Multisource, Multimodal, and Multilingual Fusion}

\author{
    \textbf{Pranav Janjani \hspace{1cm} Mayank Palan \hspace{1cm} Sarvesh Shirude} \\
    \textbf{Ninad Shegokar\hspace{1cm} Sunny Kumar \hspace{1cm} Faruk Kazi} \\
    Centre of Excellence in Complex and Nonlinear Dynamical Systems, VJTI \\
    Mumbai, India \\
    \texttt{prjanjani\_b21@ce.vjti.ac.in, mbpalan\_b22@it.vjti.ac.in, seshirude\_b20@el.vjti.ac.in,} \\ 
    \texttt{nsshegokar\_b23@ee.vjti.ac.in,skumar\_p21@ee.vjti.ac.in, fskazi@el.vjti.ac.in}
}

\begin{document}
\maketitle
\begin{abstract}
 Recent advances in large language models (LLMs) have led to new summarization strategies, offering an extensive toolkit for extracting important information. However, these approaches are frequently limited by their reliance on isolated sources of data. The amount of information that can be gathered is limited and covers a smaller range of themes, which introduces the possibility of falsified content and limited support for multilingual and multimodal data. The paper proposes a novel approach to summarization that tackles such challenges by utilizing the strength of multiple sources to deliver a more exhaustive and informative understanding of intricate topics. The research progresses beyond conventional, unimodal sources such as text documents and integrates a more diverse range of data, including YouTube playlists, pre-prints, and Wikipedia pages. The aforementioned varied sources are then converted into a unified textual representation, enabling a more holistic analysis. This multifaceted approach to summary generation empowers us to extract pertinent information from a wider array of sources. The primary tenet of this approach is to maximize information gain while minimizing information overlap and maintaining a high level of informativeness, which encourages the generation of highly coherent summaries.\end{abstract}

\section{Introduction}
Within the contemporary paradigm of omnipresent information access, where knowledge dissemination transcends traditional boundaries of format and language, the ability to efficiently extract and synthesize significant volumes of data becomes paramount. Methods have been implemented for textual summarization on PDFs from sources like educational documents and scientific research papers \cite{allahyari2017text, liu2019text}. For research, there are significant amounts of unused information in the form of a large number of scholarly articles on websites like arXiv \cite{jiang2024bridging,IBRAHIMALTMAMI20221011}. An intelligent paper search module is integrated that works just like a seasoned navigator, which searches for appropriate research papers according to the query entered by the user. This focused approach mitigates the need to go through a plethora of irrelevant information. Once research paper harvesting is complete, latest summary methodology is employed to collate the retrieved papers into short, clear, and informative summaries \cite{an2021retrievalsum}. This way, researchers and students can quickly investigate the latest developments in their area of interest without expending too much time and effort.

YouTube is another source of information as it is one of the most popular websites characterized by abundant educative and entertaining information. The process employed is more advanced than just textual analysis, as the method also generate transcripts from multilingual videos \cite{lin2024videoxum, 10077016}. Advance methods in extracting frames to capture the visual narrative embedded in the film are employed \cite{article}. By dissecting what is being said, one can better understand the topic. Therefore, the keyframes and generated transcripts are used to produce the summary.

To further improve the context of the information retrieved, summaries from reliable knowledge base such as Google, DuckDuckGo, and Wikipedia are concatenated \cite{inproceedingss, nemoto2021tool}. These resources are vast libraries of filtered, verified human knowledge and provide essential background information. By integrating information from the sources to which it is connected, the system thus manages a more profound understanding of the underlying relationships between concepts and entities. A summary of a scientific result is extracted from these vast information banks that not only describes the result but also embeds it in the appropriate historical context and the framework of scientific ideas. This well-defined structure allows a deep understanding of the content, enabling the easy integration of information from several modalities and languages into the document \cite{yuan2019knowledge}. The ultimate summary, therefore, provides a comprehensive guide that contains all the relevant information related to YouTube videos, scholarly papers, and background sources.

This paper presents a methodology for multi-source summarization, capable of gleaning nuanced details and insights from a confluence of heterogeneous data streams – multilingual, multimodal, and encompassing the full spectrum of textual, visual, and conceptual information. This holistic approach fosters a cohesive understanding by leveraging the complementary strengths of diverse information sources, ensuring a comprehensive and multifaceted representation of the underlying knowledge. This will enhance user comprehension and make navigating the ever-increasing knowledge landscape smoother by combining the power of several sources into one joint knowledge base.

% Imagine a synthesis that threads the research report, relevant scientific findings, and historical background into one cohesive narrative. The novel approach will, in collecting and generating summaries from multisource, multimodal, and multilingual data, profoundly change our relationship with information at a fundamental level.

\section{Related Work}
The wealth of multimodal and multilingual information demands valuable ways of summarizing and synthesizing information with minimal overlap. Recent works have released a multilingual abstractive summarization test set using a large dataset with multiple languages, human summaries, and pairs of the source document and its summary in different languages \cite{cao2020multisumm, hasan2021xlsum}. This milestone work provides evidence of an opportunity for automatic systems to synthesize culturally appropriate and coherent summaries. Based on such principles, video summarization frameworks have appeared with the addition of transcription and thematic analysis of video content.For instance, a documentary synopsis is supposed to include important visuals and summarize the major events and characters covered \cite{otani2016video}. All these go a step further than merely summarizing text, considering audiovisual aspects in making a holistic narrative \cite{zhou2018deep, lin2024videoxum}. Providing summaries of videos that focus on the key visual and thematic elements have also been outlined \cite{gianluigi2006innovative}.

Studies have pointed out that scientific discoveries need to be placed into relevant cultural and historical contexts, which enhances understanding and intercultural communication. Therefore it, underscores the value of using background information from comprehensive knowledge bases such as Google, Wikipedia and DuckDuckGo.%\cite{inproceedingss} 

There is a need for reinforcing summarization techniques by utilizing advanced algorithms. Some techniques emphasized retrieval-based mechanisms to improve the relevance and informativeness characteristics of summaries \cite{liu2024robust, syed2023citancecontextualized}. Summarizing research papers calls for specialized models and domain adaptation techniques to deal with a broad scope of domain-specific challenges, including scientific terms and complex syntactic structures. An attractive approach for summarizing scientific documents through joint fact detection in citations is proposed which identifies noun phrases from citation sentences that frequently co-occur. It represent common facts discussed from different perspectives. These facts could then be compiled using this multi-document summarization system into a comprehensive summary \cite{CHEN2014246}. Another literature investigates multi-document summarization techniques that evaluate various models, including graph-based, neural network-based, and hybrid approaches, proposing methods to effectively synthesize information from multiple papers \cite{MultiSci}.

The prevailing information extraction and summarization methodologies are predominantly characterized by their singular source dependence and a dearth of multi-modality. This constraint restricts the potential for optimal knowledge acquisition. This singular source dependence often leads to redundancy within extracted information and hinders the capture of diverse perspectives. To achieve a more comprehensive understanding of a subject, it is imperative to leverage information from a multiplicity of sources.To address these limitations, research efforts should be directed towards the development of robust multi-source information extraction and summarization techniques. 

\begin{figure*}[htbp]
    \centering
    \includegraphics[width=1\textwidth]{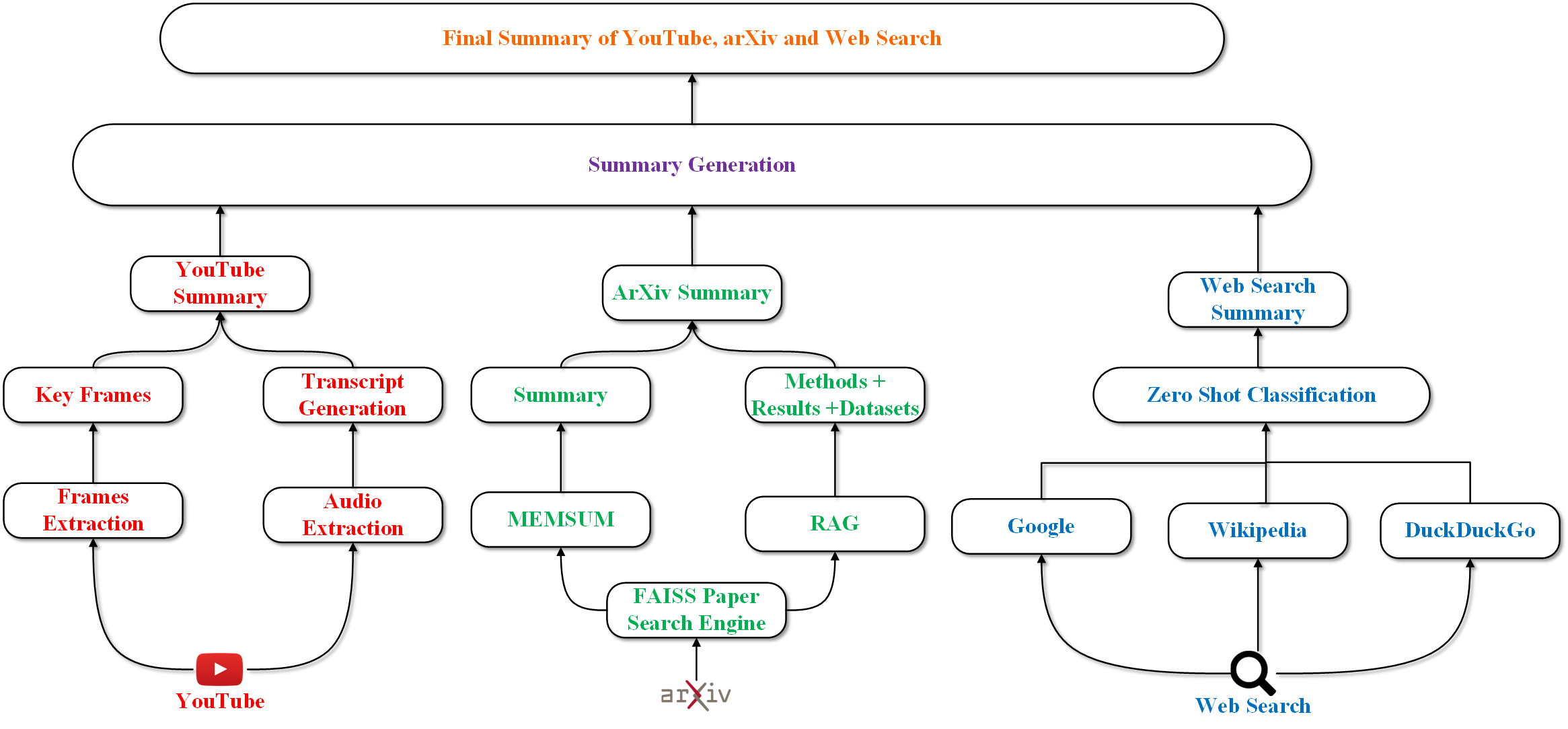}
    \caption{Methodology}
    \label{fig:wide_image}
    
\end{figure*}

The proposed multifaceted approach should serve a dual purpose: it not only mitigates redundancy within the extracted data but also promotes the inclusion of diverse and potentially conflicting perspectives. This comprehensive methodology should enhance the overall quality of the data by optimizing its relevance and breadth. Minimizing repetitive information, ensures the capture of a broader spectrum of perspectives and insights, thereby enriching the dataset with a nuanced and multifarious understanding of the subject matter.

\section{Methodology}
This paper proposes a multisource, multimodal, and multilingual information extraction system, the first of its kind to capture the most important and diverse information to reduce hallucinations and increase the quality of summary generation. Figure~\ref{fig:wide_image}, provides a comprehensive overview of the process. The functions and methods are categorized into these parts:
\begin{enumerate}
    \item Information Conversion
    \item Information Search \& Retrieval
    \item Information Convergence
\end{enumerate}

Three multimodal sources- YouTube Playlists, arXiv Papers and Web Search are considered. Each is passed through the mentioned processes to generate a superior system capable of providing robust information related to any subject matter.

\subsection {YouTube Playlists: Multilingual and Multimodal Approach}
Within the dynamic domain of information extraction, where the primary objective revolves around the harvesting of valuable knowledge from disparate sources, YouTube playlists emerge as an intriguing and relatively unexplored research frontier. By incorporating information gleaned from playlists in conjunction with established textual sources, a system can cultivate a more comprehensive and richly multifaceted comprehension of the target domain \cite{1390999}.

\subsubsection{Information Conversion}

YouTube encompasses multimodal information-audio and video each processed separately and merged to generate an optimal information representation of the query presented by the user.

(i) Audio Extraction:
The audio stream gets extracted using the YouTube API and the audio is subsequently converted to text along with its timestamp using Open AI’s robust speech recognition model Whisper \cite{10.5555/3618408.3619590}. It can identify the language of the video and transcribe the audio to a desired language. Whisper is a Transformer sequence-to-sequence model trained on various speech processing tasks, including multilingual speech recognition, speech translation,spoken language identification, and voice activity detection. An alternate method of using YouTube's Closed Caption generation was explored and provided poor results due to the lesser accuracy of the captions generated and limited multilingualism. The mentioned support is provided only for a smaller percentage of the videos so it's not optimal, adaptable or scalable.

\begin{figure}[htbp]
    \centering
    \includegraphics[width=1\linewidth]{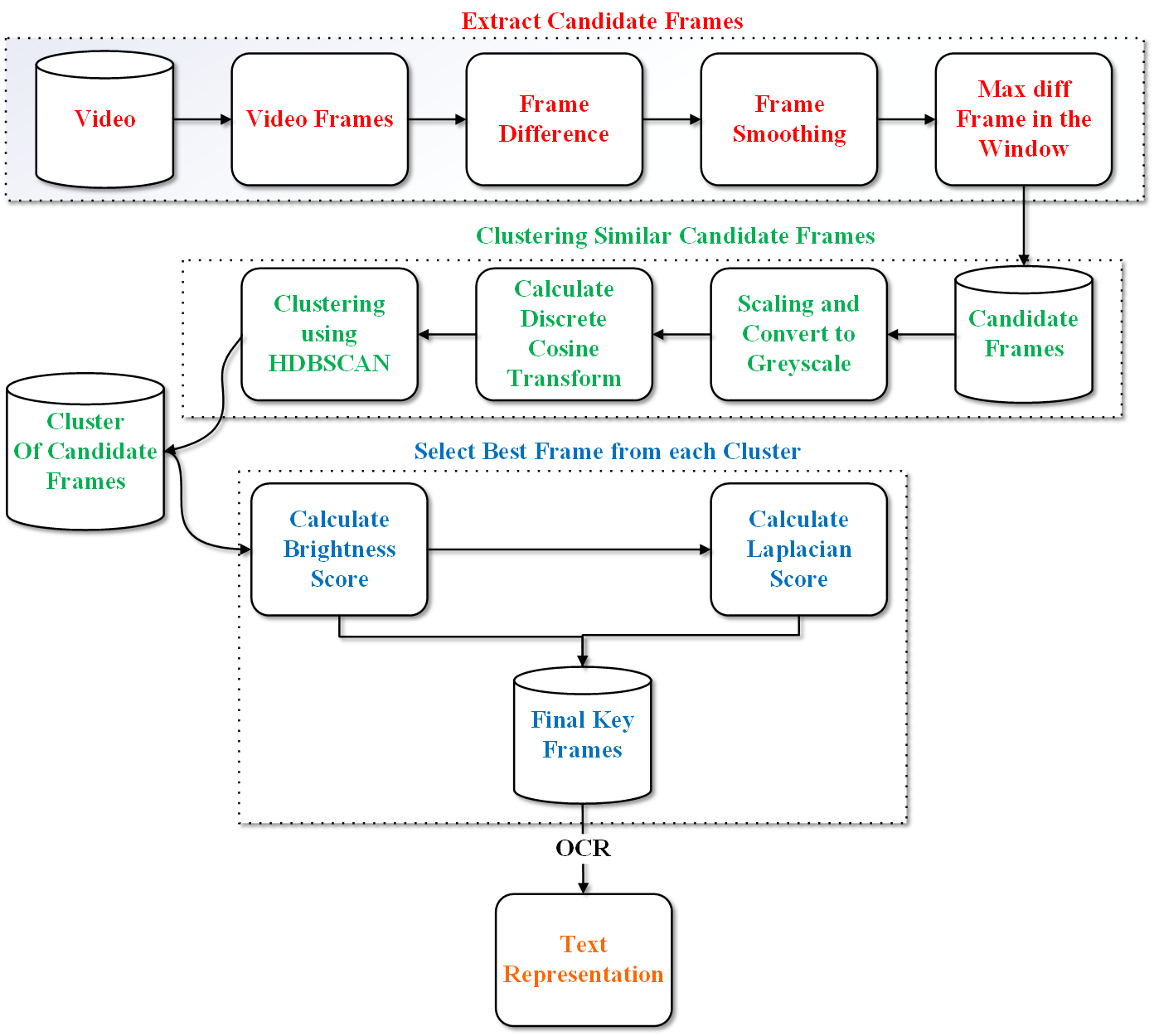}
    \caption{Key frames and Information Extraction}
    \label{fig:frames}
\end{figure}

(ii) Video Frames Extraction: 
The process begins with extracting frames from the video, followed by calculating frame differences, identifying local maxima, and selecting high-quality frames through clustering \cite{karticle, 1000355}. The resulting keyframes are processed using OCR to extract the text for a comprehensive inclusion of information. Figure \ref{fig:frames} illustrates frame extraction.
\\
To identify frames that exhibit significant changes, we calculate the sum of absolute differences (SAD) between consecutive frames. This process involves several steps:
\begin{itemize}
    \item Conversion to HSV: Each frame is converted to the HSV color space, which provides better differentiation of visual information compared to the RGB space.
    \item Difference Calculation: The absolute difference between the current frame and the previous frame is computed.
    \item Sum of Differences: The differences are summed to obtain a single value representing the change between frames.
\end{itemize}
A custom Frame class stores each frame along with its computed SAD value. These values help identify frames with substantial content changes.To extract frames that are significantly different from their neighbors, we use local maxima detection. This involves:
\begin{itemize}
    \item Smoothing: The frame difference values are smoothed using a window function (e.g., Hanning window).
    \item Local Maxima Identification: Local maxima within the smoothed difference values are identified using scipy.signal.argrelextrema.
\end{itemize}
Frames corresponding to these local maxima are considered candidate keyframes, representing moments of significant visual change in the video.
Once candidate keyframes are extracted, they undergo further evaluation based on brightness, entropy, and sharpness to ensure high visual quality:
\begin{itemize}
    \item Brightness Score: This is computed as the average brightness of the frame in the HSV color space.
    \item Entropy Score: This measures the amount of information or randomness in the frame, calculated using a disk-shaped structuring element.
    \item Sharpness (Laplacian Variance): This assesses the sharpness of the frame by computing the variance of the Laplacian, with higher values indicating sharper images.
\end{itemize}
Frames with optimal brightness and contrast (entropy) are filtered for further analysis.To ensure diversity among the selected keyframes, clustering techniques are employed:
\begin{itemize}
    \item Discrete Cosine Transform (DCT): Each candidate keyframe is converted to grayscale and resized to a standard dimension. The DCT is then applied to capture frequency components, which are used as feature vectors.
    \item HDBSCAN Clustering: The feature vectors are clustered using HDBSCAN, which groups frames into clusters based on similarity.
    \item Best Frame Selection: Within each cluster, the frame with the highest Laplacian variance is selected as the best representative frame. This step ensures that the final set of key frames is sharp and visually distinct.
\end{itemize}
Following the initial feature extraction stage, where key frames are isolated and synchronized with corresponding timestamps, the visual data undergoes text recognition and extraction. Google's Generative Pre-trained Transformer, also known as Gemini, is utilized for this purpose of Optical Character Recognition. The resulting time-stamped textual elements, gleaned from the processed video frames, are subsequently subjected to a data fusion process. This entails the seamless integration of the extracted time-stamped text with the existing, synchronized transcript. This enriched dataset, encompassing both visual and textual information with precise temporal alignment, serves as the foundation for subsequent information convergence process.

\subsubsection{Information Search and Retrieval}

Open-sourced Playlist search libraries are utilized to facilitate real-time information retrieval. The limitations of static datasets are eschewed in favor of a dynamic approach that incorporates the ever-evolving corpus of information on YouTube \cite{abuelhaija2016youtube8m}. This approach facilitates the retrieval of a vast volume of playlists, which are then be subjected to an optimal information conversion process. By simple entry of keywords as queries, the libraries retrieve the relevant playlists.

\subsubsection{Information Convergence}
To efficiently extract key information from video playlists, a synopsis generation paradigm is employed. This approach entails processing each video within the playlist to produce a concise summary that retains crucial details while eliminating redundant content. To tailor summaries to the specific content domain of the playlist, the LLaMA3 70b model's power is coupled with meticulously crafted prompt engineering techniques. Through this synergistic approach, it is ensured that the generated summaries accurately capture the diverse content of each video and the overarching narrative of the playlist.

\subsection{arXiv Paper Search and Summarization:}
The inclusion of arXiv and research papers is quite beneficial and indispensable. They offer a unique window into the frontiers of scientific knowledge, providing access to cutting-edge discoveries, detailed exploration of all related topics, and a structured format that facilitates accurate information extraction. By incorporating these valuable resources, the system can gain a deeper understanding of the information landscape and extract more comprehensive, nuanced, and insightful knowledge.
\subsubsection{Information Conversion}
The arXiv dataset is a repository of 1.7 million articles \cite{arXiv_org_submitters_2024}, with relevant features such as article id, article authors, titles, abstract, categories, full text PDFs, and more. The title and abstract for each paper were combined and converted to vector embedding using sentence transformers ‘all-MiniLM-L6-v2’ model. It maps sentences \& paragraphs to a 384-dimensional dense vector space and can be used for various natural language tasks like semantic search \cite{10.5555/3495724.3496209}.

\subsubsection{Information Search and Retrieval}

Retrieval Augmented Generation (RAG) system using Facebook AI Similarity Search (FAISS) was built to search for research papers in the arXiv dataset based on the user query and retrieves the top ‘n’ papers \cite{10.5555/3495724.3496517}. These embedded vectors are stored in the RAG System using a FAISS index, which is an open-source library designed to quickly search for embeddings in large datasets and provides scalable similarity search functions \cite{douze2024faiss}.

Upon submission of a user search query, the query text will be vectorized. These query vectors will then be compared to the vectors stored in the FAISS index. Further re-ranking is using the 'cross-encoder/ms-marco-MiniLM-L-6-v2' encoder model to improve vector search results as it compares text instead of vectors. Overall, it reorders the text search result based on the search query by assigning a score to the text result. The re-ranked search results are returned which consists of paper title, abstract, its categories, and a link to the PDF.Text extraction techniques are employed to convert the document, located at a specific URL, into a machine-readable format. 

\subsubsection{Information Convergence}
Subsequently, the summarization engine consists of two distinct levels to ensure the capture of all critical information. The first level focuses on the faithful extraction of statistical and mathematical expressions within the paper while the second level delves into the background knowledge presented and the novel findings introduced by the research, guaranteeing an inclusive and informative summary. This ensures the inclusion of crucial statistical and mathematical equations, while simultaneously retaining the significance of background knowledge and novel findings presented within the research paper.

(i) MeMSum Pipeline:
The extracted text is summarized using Multi-step Episodic Markov decision process extractive Summarizer (MeMSum). It is a reinforcement learning-based extractive summarizer enriched at each step with information on the current extraction history. MemSum's ‘mesum-arXiv-summarization’ pre-trained model is used to summarize the research papers \cite{gu-etal-2022-memsum}. 

\begin{figure}
    \centering
    \includegraphics[width=1\linewidth]{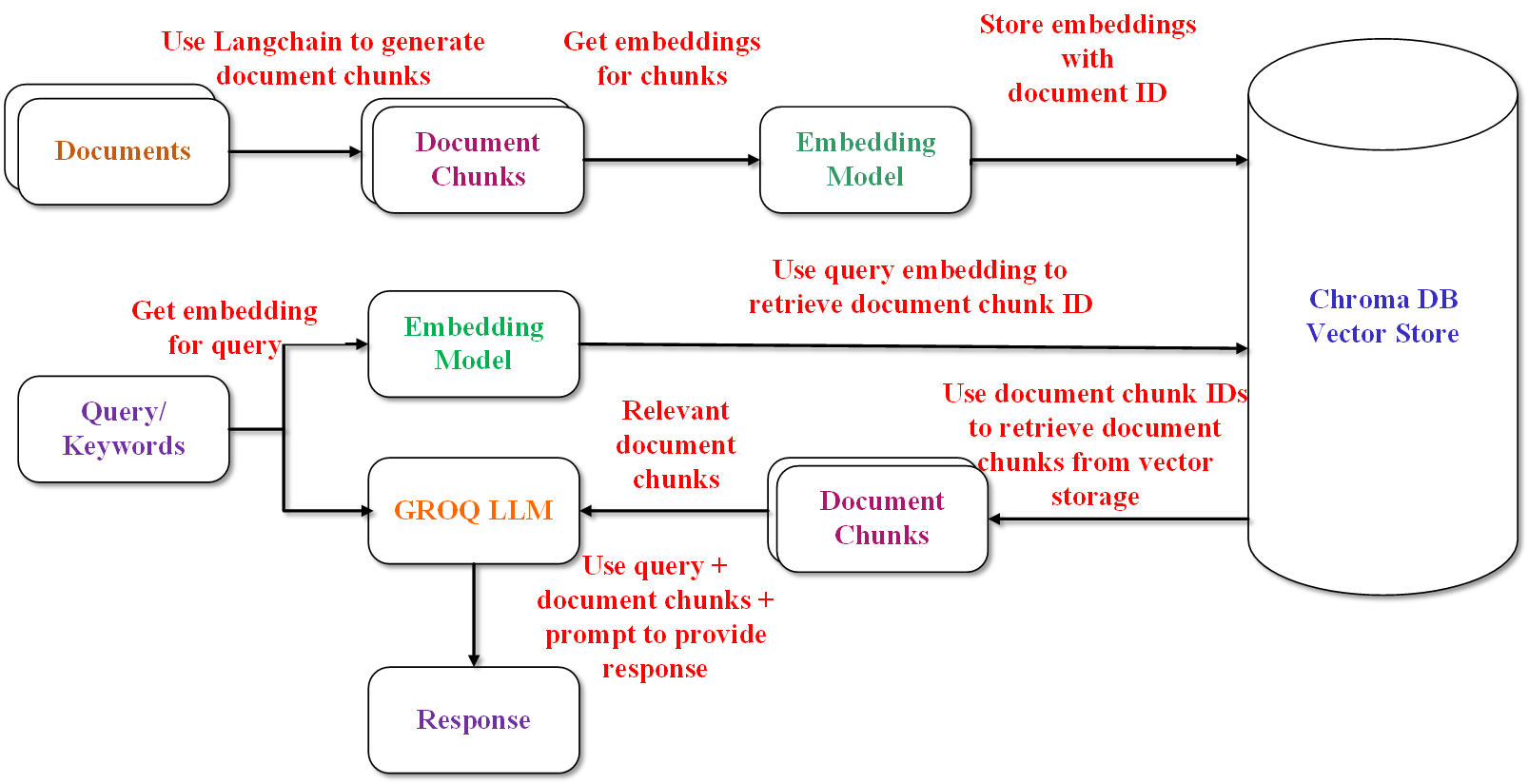}
    \caption{RAG Chain Pipeline}
    \label{fig:rag}
\end{figure}
(ii) RAG Chain Pipeline:
The research paper’s information is segmented into manageable chunks using Langchain. Each chunk is then transformed into a numerical representation, or embedding, via an embedding model, which is ‘BAAI/bge-small-en-v1.5’ capturing semantic nuances necessary for efficient retrieval \cite{chen2024bge}. These embeddings, alongside document IDs, are stored in a Chroma DB Vector Store, enabling quick, similarity-based retrieval. This storage mechanism ensures that the document chunks can be easily accessed and utilized in subsequent stages of the pipeline. Figure~\ref{fig:rag} illustrated the mechanisms of the RAG chain pipeline.

Upon receiving a query, an embedding model generates a corresponding query embedding, which is utilized to fetch relevant document chunks from the Chroma DB Vector Store. These chunks, containing pertinent information, are fed into a GROQ LLM, which processes the query and contextually integrates the document chunks. The LLM is prompted to retrieve and synthesize information specifically about the methodology, datasets, and results of the research papers. This approach maximizes the utility of large language models by augmenting them with specific, contextual data retrieved efficiently through embeddings and vector stores, ensuring the responses are accurate, relevant, and comprehensive.

Following the independent processing by the MeMSum and RAG Chain pipelines, their outputs are synergistically merged to create a holistic textual representation. The resulting comprehensive nature of this information source facilitates the efficacious dissemination of knowledge. This ensures that users, particularly those engaged in academic pursuits (i.e., students and researchers), possess a contemporaneous understanding of the latest developments within their respective fields. Furthermore, this comprehensive approach bolsters the source's epistemic authority by establishing a foundation for information retrieval that is demonstrably robust. Consequently, the likelihood of encountering unsubstantiated assertions or misleading data or hallucinations is significantly reduced.

\subsection{Web Search Engine}
To mitigate the risk of data obsolescence and guarantee exhaustive information retrieval from a diverse range of sources, a multifaceted web search strategy is employed. This strategy leverages the capabilities of prominent search engines, namely Wikipedia, DuckDuckGo, and Google.
\subsubsection{Information Search and Retrieval}
 The aforementioned search engines function as established web crawlers, systematically traversing the web to index content provided as "llama-index" agents. Extracted webpages undergo a rigorous topic classification process using a large language model (LLM) based Zero-Shot classification technique \cite{yin2019benchmarking, puri2019zeroshot}. This classification ensures that only those webpages demonstrably aligned with the research query's thematic domain are retained for subsequent analysis.

\subsubsection{Information Convergence}
Our approach involves synthesizing a vast array of pertinent data retrieved from numerous web search engines. Key insights are distilled from this extensive collection, significantly reducing redundant information and ensuring the content remains current and relevant. This is accomplished by harnessing the formidable capabilities of the LLaMA3 70b model, expertly combined with meticulously crafted prompt engineering techniques \cite{llama3modelcard}. This methodology guarantees that the generated summaries not only encapsulate the essence of each source but also cohesively convey the overarching narrative of the entire topic and its related concepts.

\subsection{Final Summary Generation}
Following the derivation of succinct representations from each source, MeMSum in a multi-source fashion is used to strategically mitigate the redundancy and overlap of information across the summaries. This approach ensures the inclusion of all salient and distinct viewpoints, fostering the generation of a comprehensive information extraction summary that encompasses multi-source, multi-modal, and multilingual content.

\section{Results}

\begin{table}[htbp]
\centering
\begin{tabular}{|l|r|}
\hline
\multicolumn{1}{|c|}{\textbf{Keyword}} & \multicolumn{1}{c|}{\textbf{Coherence Score}} \\ \hline
\textbf{Deep Learning}                 & 0.470                                         \\ \hline
\textbf{Statistics}                         & 0.462                                         \\ \hline
\textbf{Quantum Physics}                       & 0.470                                         \\ \hline
\end{tabular}
\caption{Comparison of Coherence Score}
\label{tab:my-table}
\end{table}

\begin{table}[htbp]
\centering
\resizebox{\columnwidth}{!}{%
\begin{tabular}{|l|r|r|r|r|}
\hline
\textbf{Metric}                        & \multicolumn{1}{l|}{\textbf{Final Summary}} & \multicolumn{1}{l|}{\textbf{arXiv}} & \multicolumn{1}{l|}{\textbf{Web Search}} & \multicolumn{1}{l|}{\textbf{YouTube}} \\ \hline
\textbf{KL Divergence (vs arXiv)}      & 0.465                                       & \multicolumn{1}{c|}{-}              & 4.791                                    & 4.431                                 \\ \hline
\textbf{KL Divergence (vs Web Search)} & 2.360                                       & 5.424                               & \multicolumn{1}{c|}{-}                   & 5.057                                 \\ \hline
\textbf{KL Divergence (vs YouTube)}    & 1.404                                       & 4.861                               & 4.808                                    & \multicolumn{1}{c|}{-}                \\ \hline
\textbf{Entropy}                       & 8.113                                       & 7.540                               & 7.993                                    & 7.739                                 \\ \hline
\textbf{TTR}              & 0.186                                       & 0.218                               & 0.254                                    & 0.134                                 \\ \hline
\textbf{Redundancy Score}              & 0.084                                       & 0.453                               & 0.578                                    & 0.235                                 \\ \hline
\end{tabular}%
}
\caption{Metrics Evaluation for `Deep Learning' sample}
\label{tab:table2}
\end{table}

\begin{table}[htbp]
\centering
\resizebox{\columnwidth}{!}{%
\begin{tabular}{|l|r|r|r|r|}
\hline
\textbf{Metric}                        & \multicolumn{1}{l|}{\textbf{Final Summary}} & \multicolumn{1}{l|}{\textbf{arXiv}} & \multicolumn{1}{l|}{\textbf{Web Search}} & \multicolumn{1}{l|}{\textbf{YouTube}} \\ \hline
\textbf{KL Divergence (vs arXiv)}      & 1.824                                       & \multicolumn{1}{c|}{-}              & 4.453                                    & 3.599                                 \\ \hline
\textbf{KL Divergence (vs Web Search)} & 0.442                                       & 4.867                               & \multicolumn{1}{c|}{-}                   & 2.636                                 \\ \hline
\textbf{KL Divergence (vs YouTube)}    & 0.925                                       & 5.141                               & 4.118                                    & \multicolumn{1}{c|}{-}                \\ \hline
\textbf{Entropy}                       & 8.491                                       & 7.998                               & 7.816                                    & 8.576                                 \\ \hline
\textbf{TTR}              & 0.133                                       & 0.289                               & 0.207                                    & 0.109                                 \\ \hline
\textbf{Redundancy Score}              & 0.055                                       & 0.810                               & 0.521                                    & 0.088                                 \\ \hline
\end{tabular}%
}
\caption{Metrics Evaluation for `Statistics' sample}
\label{tab:table3}
\end{table}

\begin{table}[!htbp]
\centering
\resizebox{\columnwidth}{!}{%
\begin{tabular}{|l|r|r|r|r|}
\hline
\textbf{Metric}                        & \multicolumn{1}{l|}{\textbf{Final Summary}} & \multicolumn{1}{l|}{\textbf{arXiv}} & \multicolumn{1}{l|}{\textbf{Web Search}} & \multicolumn{1}{l|}{\textbf{YouTube}} \\ \hline
\textbf{KL Divergence (vs arXiv)}      & 2.250                                       & \multicolumn{1}{c|}{-}              & 5.252                                    & 4.350                                 \\ \hline
\textbf{KL Divergence (vs Web Search)} & 0.449                                       & 4.900                               & \multicolumn{1}{c|}{-}                   & 2.524                                 \\ \hline
\textbf{KL Divergence (vs YouTube)}    & 0.704                                       & 4.531                               & 3.442                                    & \multicolumn{1}{c|}{-}                \\ \hline
\textbf{Entropy}                       & 8.348                                       & 8.072                               & 7.925                                    & 8.203                                 \\ \hline
\textbf{TTR}              & 0.158                                       & 0.317                               & 0.255                                    & 0.148                                 \\ \hline
\textbf{Redundancy Score}              & 0.038                                       & 0.693                               & 0.387                                    & 0.114                                 \\ \hline
\end{tabular}%
}
\caption{Metrics Evaluation for `Quantum Physics' sample}
\label{tab:table4}
\end{table}

\begin{table*}[ht]
\centering
\resizebox{\textwidth}{!}{%
\begin{tabular}{lccccccccc}
\hline
\multicolumn{1}{c}{\multirow{2}{*}{\textbf{Category}}} & \multicolumn{3}{c}{\textbf{Rouge-1}}               & \multicolumn{3}{c}{\textbf{Rouge-2}}               & \multicolumn{3}{c}{\textbf{Rouge-L}}               \\ \cline{2-10} 
\multicolumn{1}{c}{}                                   & \textbf{Recall} & \textbf{Precision} & \textbf{F1} & \textbf{Recall} & \textbf{Precision} & \textbf{F1} & \textbf{Recall} & \textbf{Precision} & \textbf{F1} \\ \hline
\textbf{Final Summary vs arXiv}                        & 0.955           & 0.446              & 0.608       & 0.941           & 0.373              & 0.534       & 0.954           & 0.446              & 0.608       \\ \hline
\textbf{Final Summary vs Web Search}                   & 0.747           & 0.381              & 0.505       & 0.702           & 0.273              & 0.394       & 0.745           & 0.380              & 0.504       \\ \hline
\textbf{Final Summary vs YouTube}                      & 0.553           & 0.505              & 0.528       & 0.411           & 0.442              & 0.426       & 0.545           & 0.498              & 0.520       \\ \hline
\textbf{arXiv vs Web Search}                           & 0.233           & 0.255              & 0.243       & 0.062           & 0.061              & 0.062       & 0.208           & 0.227              & 0.217       \\ \hline
\textbf{arXiv vs YouTube}                              & 0.177           & 0.345              & 0.234       & 0.046           & 0.125              & 0.068       & 0.166           & 0.324              & 0.219       \\ \hline
\textbf{Web Search vs YouTube}                         & 0.161           & 0.288              & 0.206       & 0.031           & 0.086              & 0.046       & 0.145           & 0.259              & 0.186       \\ \hline
\end{tabular}%
}
\caption{ROUGE Score metric for `Deep Learning' sample}
\label{tab:table5}
\end{table*}

\begin{table*}[ht]
\centering
\resizebox{\textwidth}{!}{%
\begin{tabular}{lccccccccc}
\hline
\multicolumn{1}{c}{\multirow{2}{*}{\textbf{Category}}} & \multicolumn{3}{c}{\textbf{Rouge-1}}               & \multicolumn{3}{c}{\textbf{Rouge-2}}               & \multicolumn{3}{c}{\textbf{Rouge-L}}               \\ \cline{2-10} 
\multicolumn{1}{c}{}                                   & \textbf{Recall} & \textbf{Precision} & \textbf{F1} & \textbf{Recall} & \textbf{Precision} & \textbf{F1} & \textbf{Recall} & \textbf{Precision} & \textbf{F1} \\ \hline
\textbf{Final Summary vs arXiv}                        & 0.714           & 0.204              & 0.317       & 0.553           & 0.107              & 0.179       & 0.693           & 0.198              & 0.308       \\ \hline
\textbf{Final Summary vs Web Search}                   & 1.000           & 0.380              & 0.551       & 1.000           & 0.293              & 0.453       & 1.000           & 0.380              & 0.551       \\ \hline
\textbf{Final Summary vs YouTube}                      & 0.622           & 0.719              & 0.667       & 0.537           & 0.682              & 0.601       & 0.617           & 0.713              & 0.661       \\ \hline
\textbf{arXiv vs Web Search}                           & 0.216           & 0.288              & 0.247       & 0.051           & 0.078              & 0.062       & 0.195           & 0.259              & 0.222       \\ \hline
\textbf{arXiv vs YouTube}                              & 0.108           & 0.438              & 0.174       & 0.026           & 0.168              & 0.044       & 0.101           & 0.408              & 0.161       \\ \hline
\textbf{Web Search vs YouTube}                         & 0.150           & 0.456              & 0.226       & 0.040           & 0.175              & 0.066       & 0.143           & 0.434              & 0.215       \\ \hline
\end{tabular}%
}
\caption{ROUGE Score metric for `Statistics' sample}
\label{tab:table6}
\end{table*}

\begin{table*}[ht]
\centering
\resizebox{\textwidth}{!}{%
\begin{tabular}{lccccccccc}
\hline
\multicolumn{1}{c}{}                                    & \multicolumn{3}{c}{\textbf{Rouge-1}}                                                          & \multicolumn{3}{c}{\textbf{Rouge-2}}                                                          & \multicolumn{3}{c}{\textbf{Rouge-L}}                                                          \\ \cline{2-10} 
\multicolumn{1}{c}{\multirow{-2}{*}{\textbf{Category}}} & \textbf{Recall}               & \textbf{Precision}            & \textbf{F1}                   & \textbf{Recall}               & \textbf{Precision}            & \textbf{F1}                   & \textbf{Recall}               & \textbf{Precision}            & \textbf{F1}                   \\ \hline
\textbf{Final Summary vs arXiv}                         & \cellcolor[HTML]{FFFFFF}0.696 & \cellcolor[HTML]{FFFFFF}0.276 & \cellcolor[HTML]{FFFFFF}0.395 & \cellcolor[HTML]{FFFFFF}0.577 & \cellcolor[HTML]{FFFFFF}0.152 & \cellcolor[HTML]{FFFFFF}0.240 & \cellcolor[HTML]{FFFFFF}0.683 & \cellcolor[HTML]{FFFFFF}0.271 & \cellcolor[HTML]{FFFFFF}0.388 \\ \hline
\textbf{Final Summary vs Web Search}                    & \cellcolor[HTML]{FFFFFF}0.968 & \cellcolor[HTML]{FFFFFF}0.397 & \cellcolor[HTML]{FFFFFF}0.563 & \cellcolor[HTML]{FFFFFF}0.956 & \cellcolor[HTML]{FFFFFF}0.306 & \cellcolor[HTML]{FFFFFF}0.464 & \cellcolor[HTML]{FFFFFF}0.968 & \cellcolor[HTML]{FFFFFF}0.397 & \cellcolor[HTML]{FFFFFF}0.563 \\ \hline
\textbf{Final Summary vs YouTube}                       & \cellcolor[HTML]{FFFFFF}0.774 & \cellcolor[HTML]{FFFFFF}0.693 & \cellcolor[HTML]{FFFFFF}0.731 & \cellcolor[HTML]{FFFFFF}0.711 & \cellcolor[HTML]{FFFFFF}0.651 & \cellcolor[HTML]{FFFFFF}0.679 & \cellcolor[HTML]{FFFFFF}0.766 & \cellcolor[HTML]{FFFFFF}0.686 & \cellcolor[HTML]{FFFFFF}0.724 \\ \hline
\textbf{arXiv vs Web Search}                            & \cellcolor[HTML]{FFFFFF}0.232 & \cellcolor[HTML]{FFFFFF}0.241 & \cellcolor[HTML]{FFFFFF}0.236 & \cellcolor[HTML]{FFFFFF}0.058 & \cellcolor[HTML]{FFFFFF}0.070 & \cellcolor[HTML]{FFFFFF}0.064 & \cellcolor[HTML]{FFFFFF}0.206 & \cellcolor[HTML]{FFFFFF}0.213 & \cellcolor[HTML]{FFFFFF}0.210 \\ \hline
\textbf{arXiv vs YouTube}                               & \cellcolor[HTML]{FFFFFF}0.171 & \cellcolor[HTML]{FFFFFF}0.386 & \cellcolor[HTML]{FFFFFF}0.237 & \cellcolor[HTML]{FFFFFF}0.040 & \cellcolor[HTML]{FFFFFF}0.138 & \cellcolor[HTML]{FFFFFF}0.062 & \cellcolor[HTML]{FFFFFF}0.156 & \cellcolor[HTML]{FFFFFF}0.353 & \cellcolor[HTML]{FFFFFF}0.217 \\ \hline
\textbf{Web Search vs YouTube}                          & \cellcolor[HTML]{FFFFFF}0.229 & \cellcolor[HTML]{FFFFFF}0.499 & \cellcolor[HTML]{FFFFFF}0.313 & \cellcolor[HTML]{FFFFFF}0.078 & \cellcolor[HTML]{FFFFFF}0.224 & \cellcolor[HTML]{FFFFFF}0.116 & \cellcolor[HTML]{FFFFFF}0.209 & \cellcolor[HTML]{FFFFFF}0.456 & \cellcolor[HTML]{FFFFFF}0.286 \\ \hline
\end{tabular}%
}
\caption{ROUGE Score metric for `Quantum Physics' sample}
\label{tab:table7}
\end{table*}

This research presents a novel methodology for gleaning information from a confluence of disparate sources. The following robust metrics are utilized to rigorously evaluate the efficacy of our information extraction and summarization system:
\subsection{Entropy}Entropy in textual summaries measures the average unpredictability or information richness of content. This dimension shows the coverage of vocabulary and topics gained from the text summary, i.e., overall diversity and coverage. 
\begin{equation}
    H(X) = -\sum_{i=1}^{|X|} p(x_i) \log_2(p(x_i))
\end{equation}
where, $H(X)$ represents the entropy of the document $X$,$|X|$ represents the total number of words in the document, and $p(x_i)$ represents the probability of the i-th word appearing in the document.

Higher values for entropy suggests an extensive spread of different words and concepts and leading towards an extremely informative and diverse summary. In contrast, lower entropy implies a predictable structure with less variety of unique terms, suggesting a narrower focus or repetitive content \cite{10263021}.

\subsection{KL Divergence}Kullback-Leibler (KL) Divergence measures the difference between two probability distributions of summary content. 
\begin{equation}
D_{KL}(P || Q) = \sum_{i=1}^{|X|} p(x_i) \log_2(\frac{p(x_i)}{q(x_i)})
\end{equation}
where, $D_{KL}(P || Q)$ represents the KL divergence between two probability distributions $P$ and $Q$ which represent the word distributions of two different documents or a document and its language model, $p(x_i)$ represents the probability of the i-th word under distribution P and $q(x_i)$ represents the probability of the i-th word under distribution $Q$.

In the context of textual summaries, it quantifies how much more information or content one summary adds compared with another. Small KL Divergence indicates similar content distributions for summaries, while larger values suggest more divergence and unique information—diverse content \cite{zhang2023properties}.

\subsection{Redundancy Score}Based on KL Divergence, this is a statistical measure used to establish how much novelty information a summary brings compared to the shared summary distribution derived from several sources. If the summary has a low redundancy score, it brings novel perspectives and information not found in other summaries; hence, it becomes invaluable for different views or information. Higher redundancy scores reflect more overlap or repetition in content with other summaries, whereas lower scores show less unique contribution in information.
\subsection{Average Coherence}It measures semantic consistency and flow of the summary by checking inter-sentential similarity between sentences or paragraphs; that is, how good ideas are connected and presented within the summary. Higher average coherence scores mean a summary is more well-structured and coherent, with smoother transitions among points. Lower average scores may indicate disjointed or less coherent content, where ideas may not flow as logically or seamlessly \cite{10.1162/tacl_a_00388}.

% $Average Coherence=$
% \begin{equation}
%     \frac{1}{n-w+1} \sum{i=w}^n \frac{\left( \frac{1}{w} \sum_{j=i-w}^{i-1} E_j \right) \cdot E_i}{\left\| \frac{1}{w} \sum_{j=i-w}^{i-1} E_j \right\| \|E_i\|}
% \end{equation}

\subsection{Evaluating Document Coherence Modeling}This paper examines how well different language models can evaluate document coherence. It delves into the methods used to measure semantic consistency and flow within summaries, specifically focusing on inter-sentential similarity and the logical progression of ideas \cite{10.1162/tacl_a_00388}.

\subsection{Type-Token Ratio (TTR)}

The Type-Token Ratio (TTR) is a measure of lexical diversity in a given text. It is defined as the ratio of the number of unique words (types) to the total number of words (tokens) in the text.

\begin{equation}
   {TTR} = \frac{V}{N}
\end{equation}
where, $V$ represents the number of unique words (types) in the text and $N$ is the total number of words (tokens) in the text
\cite{abc}.

\subsection{ROUGE Scores}Recall-Oriented Understudy for Gisting Evaluation (ROUGE) scores compare summaries by measuring their efficiency at capturing information critical in reference texts. The same n-grams (words, phrases) overlapping between a generated summary and reference text show the statistics of precision, recall, and F1 measures for these.

High ROUGE scores reflect high agreement and overlap between a generated summary and its reference, being more exact or complete in that sense. Low scores indicates gaps or incongruities in the information or misunderstanding of the original content of the summary \cite{lin-2004-rouge}.

The methodology presented in the paper has been extensively evaluated over a plethora of examples. Three samples - ‘Deep Learning’, ‘Statistics’ and ‘Quantum Physics’ are presented below. The information obtained from these sources is effectively utilized by the proposed methodology.

Table \ref{tab:my-table} depicts that the final summaries are coherent, they follow a logical flow and and allow the reader to navigate the information extracted sensibly and it ensures the abruption caused due to multiple sources information integration is minimal.

Tables \ref{tab:table2}, \ref{tab:table3}, and \ref{tab:table4} contain  metrics used to evaluate the quality of the final summary compared to summaries from individual sources. These metrics include KL Divergence, Entropy, Type Token Ratio, and Redundancy Score. The KL Divergence metric suggests that the divergence between individual sources as evident is quite high, which supports our claim that the information brought in by different sources is unique in nature. Relatively, the KL divergence between Final Summary and different sources is less which shows that our final summary integrates information from different sources properly. This integration results in a summary that is diversified and rich in information content, as indicated by high entropy. It utilizes an enriched vocabulary, reflected by the type token ratio, and effective in expressing information, as shown by the low redundancy score. Consequently, information from different sources makes the description of the topic more detailed and balanced compared to what would have been achieved in any one source.

From Table \ref{tab:table5}, \ref{tab:table6} and \ref{tab:table7}  it is evident that high recall scores in comparisons with individual sources indicate that the final summary contains most relevant content from the individual sources. The moderate values of the precision scores prove that the final summary has more context or information than the individual sources. The F1 scores of the final summary are consistent, indicating a balance between recall and precision. Such a balance in the F1 scores means a final summary informed of the synthesis of information but not performed at the expense of relevance and coherency. The shallow ROUGE scores of the three different sources, arXiv, Wikipedia, and YouTube, document their particular contribution to the presented work. The high distinctiveness value in the ROUGE scores points to the worth of combining these multiple sources, particularly in making a final summary of worth both rich and detailed. In general, the hypothesis that the given topic is fully covered in this multi-channel, multimodal and multilingual setting, taking advantage of idiosyncrasy offered by each source, gets backed up by the ROUGE scores.

\section{Conclusion}
In conclusion, this paper culminated in the establishment of a robust knowledge repository through multifaceted information extraction. The polyvalent approach, defined by the meticulous elimination of redundant data and the strategic inclusion of divergent viewpoints, demonstrably enhanced the thematic relevance and extensiveness of the constructed dataset. The potency of this methodology was further underscored by a comparative evaluation of metrics across information sources, encompassing entropy, KL divergence, type token ratio, and redundancy scores. Notably, the generated summaries exhibited a minimal degree of redundancy coupled with optimal type token ratio. Additionally, the deployment of the ROUGE metric battery elucidated the intricate interplay between textual similarity and coverage within the final product and the source materials. This reinforces the significance of incorporating diverse perspectives in achieving a comprehensive understanding. Furthermore, the proposed framework exhibited exceptional coherence, bolstering its capacity to safeguard thematic pertinence and internal consistency. In simpler terms, the holistic data extraction approach employed in this research yielded a high-quality dataset. This optimized thematic relevance and the overall breadth of the information corpus by meticulously attenuating redundancy and strategically incorporating diverse perspectives. The metrics and comparative analyses presented throughout this work provide unequivocal support for the effectiveness of this strategy, further solidifying its potential to significantly impact the knowledge foundation and analytical resilience of technical research endeavors.

\section{Acknowledgments}

Authors acknowledge the support of the Centre of Excellence (CoE) in Complex \& Nonlinear Dynamical Systems (CNDS) under TEQIP-III funding.

\bibstyle{alphabetic}
\bibliography{sources}

\end{document}